\documentclass[11pt]{article}

\usepackage[margin=1in]{geometry}
\usepackage{amsmath,amssymb}
\usepackage{array}
\usepackage{booktabs}
\usepackage{enumitem}
\usepackage{float}
\usepackage{microtype}
\usepackage{multirow}
\usepackage{natbib}
\usepackage{tabularx}
\usepackage{xcolor}
\usepackage{url}
\usepackage[hidelinks]{hyperref}

\definecolor{routeblue}{HTML}{245B8A}
\definecolor{mergeorange}{HTML}{A85720}
\definecolor{softgray}{HTML}{F3F4F6}

\hypersetup{
  colorlinks=true,
  linkcolor=routeblue,
  citecolor=routeblue,
  urlcolor=routeblue,
  pdftitle={Stage-Replay Divergence Follows the KV Cache},
  pdfauthor={Alexander Boesgaard Lorup}
}

\setlist[itemize]{leftmargin=1.35em,itemsep=0.2em,topsep=0.25em}
\setlist[enumerate]{leftmargin=1.55em,itemsep=0.2em,topsep=0.25em}
\renewcommand{\arraystretch}{1.12}

\newcolumntype{Y}{>{\raggedright\arraybackslash}X}
\newcolumntype{L}[1]{>{\raggedright\arraybackslash}p{#1}}
\newcolumntype{R}[1]{>{\raggedleft\arraybackslash}p{#1}}

\title{\textbf{Stage-Replay Divergence Follows the KV Cache}\\
\large Fixed-Prefix Precision Controls and Bidirectional Cache Transplantation}

\author{Alexander Boesgaard Lorup\\
\small Openhagen}

\date{July 30, 2026}

\begin{document}
\maketitle

\begin{abstract}
Stage-replay diagnostics reconstruct intermediate token prefixes and treat
fresh-prefill continuation as continuation from the decoder state that
originally reached the prefix.  We audit that assumption at a whole reasoning-
stage boundary in a Qwen2.5-derived system.  A matched 200-item experiment
compares retained live cache with one-shot prefill of identical integer
tokens and places an exact replica on both sides.  In BF16, replicas remain
exact while the constructions differ on 166 suffixes and 20 correctness
labels; the accuracy difference is only one point (paired 95\% CI
\([-3.5,+5.5]\)).

A fixed-prefix \(2\times2\) holds all 200 token states constant while crossing
construction and precision.  The BF16 disagreements recur, whereas FP32
produces no decoded disagreement (95\% Wilson upper bound 1.88\%).  A
prospective bridge makes token-by-token incremental and retained live caches
bit-exact on 12/12 rows; an all-200 saved-ledger audit reproduces every retained
trajectory and comparison fingerprint.  Bidirectional transplantation of all
48 key/value layers makes every tested divergent continuation follow its
cache donor, both on a selected set at the primary checkpoint (24/24) and an
outcome-blind replication at a later checkpoint (43/43).  Exact-token replay
can therefore be repeatable without preserving live-state fidelity.  On the
tested states, boundary K/V cache is a causally sufficient carrier of the
divergent trajectory, while numerical precision moderates its behavioral
expression.
\end{abstract}

\section{Introduction}

Stage-replay diagnostics ask a model to continue from an intermediate point
in a previously generated trajectory.  They support counterfactual
token-credit estimation, prefix interventions, fixed-context aggregation
probes, and distillation from privileged intermediate states.  One documented
shortcut in counterfactual token-credit work reconstructs the stored text as a
fresh prompt \citep{matteson2026refeeding}.  Identical token IDs and agreement
across repeated fresh-prompt runs can then be mistaken for evidence that the
reconstructed continuation matches what the original decoder would have
produced from that point.

That inference is stronger than the checks justify.  A live autoregressive
decode reaches a boundary through incremental cache updates; fresh replay
recomputes the same prefix by prefill.  Cache-on and cache-off execution can
differ in kernel structure, memory layout, and floating-point accumulation
order \citep{chodavarapu2026kvcache}.  Batch composition can likewise alter
reduction order and outputs \citep{yuan2025nondeterminism}, while re-feeding
changes both prefix construction and step-level batch dynamics
\citep{matteson2026refeeding}.  Repeatability only shows that each execution
contract is stable.  Token identity only shows that both contracts consume
the same discrete prefix.  Neither establishes equality of the internal
state or its downstream trajectory.

We study this distinction in a Qwen2.5-derived system that generates three
reasoning branches, a learned merge, and a final answer.  The boundary after
the merge-opening token provides a whole-stage replay test.  A matched matrix
separates replica noise from live-versus-prefill disagreement; a fixed-prefix
crossing holds the integer state constant over BF16 and FP32; and a direct
intervention transplants the complete boundary K/V cache in both directions.
Every continuation uses one suffix decoder and a matched token, role, mask,
position, limit, and physical-batch contract.

The result is a cancellation-heavy fidelity failure rather than an accuracy
effect.  Replicas are exact, yet BF16 live and prefill states frequently lead
to different trajectories while their net scores remain close.  The
fixed-prefix crossing removes the discrete-state confound: the BF16 effect
recurs and no decoded disagreement is observed in FP32.  A prospective bridge
then verifies that token-by-token incremental construction reproduces an
ordinary live cache when both consume the same newly reached prefix.

Direct intervention supplies the state-level result.  On every divergent row
in both the selected primary-checkpoint set and an outcome-blind later-
checkpoint replication, the continuation follows the donor after the complete
K/V cache is swapped while recipient token state and boundary logits are
retained.  This identifies the cache as a sufficient carrier without claiming
that one particular kernel operation causes the underlying numerical drift.

The audit originated in a self-training analysis that interpreted replayed
successful branch contexts as evidence for a late merge-conversion barrier.
That interpretation assigned failed policy improvement to a particular
reasoning stage.  Once fresh-prefill replay failed the live-state control, the
conversion estimate no longer described the end-to-end pipeline.  This
invalidated claim motivates the audit; the training campaign itself is not a
contribution of this paper.

We make three contributions:

\begin{itemize}
  \item \textbf{Matched fidelity and precision controls.}  A retained-live
        matrix isolates boundary-state construction, a fixed-prefix
        \(2\times2\) identifies precision moderation, and a prospective
        bridge connects token-by-token construction to ordinary live decoding.
  \item \textbf{Direct state intervention.}  Bidirectional whole-cache
        transplantation makes every divergent suffix follow the K/V donor in
        both a selected primary-checkpoint set and an outcome-blind checkpoint
        replication, with exact replicas and nondivergent controls.
  \item \textbf{A compact fidelity protocol.}  The design separates replica
        stability, token identity, live-state fidelity, and causal state
        sufficiency while retaining task-level paired outcomes.
\end{itemize}

We do not claim a replay accuracy gain or harm, a universal BF16/FP32 law, a
universal kernel or scheduler cause, or the absence of latent stage structure.
The claim is operational: under the
tested ordinary BF16 surface, exact-token fresh prefill is not a faithful
substitute for incremental decoder state; full K/V state carries every tested
divergent suffix in the two transplantation experiments; and high precision
removes the observed behavioral difference on the tested fixed prefixes.

\section{Related Work}

\subsection{Inference nondeterminism and replay fidelity}

LLM outputs depend on the execution contract even under nominally
deterministic decoding.  Batch size, GPU count, and hardware can alter
responses and benchmark scores because early rounding differences cross
later decision boundaries \citep{yuan2025nondeterminism}.  Recent systems
address this dependence through different mechanisms: TBIK aligns reduction
orders across tensor-parallel configurations, whereas \textsc{MarginGate}
verifies low-margin BF16 steps and repairs confirmed mismatches in the current
K/V column \citep{zhang2026tbik,chu2026margingate}.

Two studies are especially close.  \citet{chodavarapu2026kvcache} compare
cache-on and cache-off inference and report deterministic FP16 trajectory
divergence that disappears behaviorally under an FP32 falsification.
Their cache-off arm recomputes the full sequence at every step and therefore
differs from our live-stage versus one-shot-prefill comparison, but it
establishes the relevant precision sensitivity.  They intervene on the
residual stream and propose direct intervention on cached K/V tensors as
future work; our experiment implements that intervention class at a
whole-stage boundary.
\citet{matteson2026refeeding} compare
continuations resumed from verified decode-time KV blocks, a second
exact-resume pass, and fresh re-feeding for counterfactual token-credit
estimation.  Re-feeding changes estimates above the replica floor; a
batch-invariant vLLM confirmation makes all stored channels identical in the
tested configuration.

We do not claim the first demonstration that re-feeding differs from state
resumption.  Our extension operates at a whole reasoning-stage boundary,
tracks long greedy trajectories and paired task correctness, fixes identical
prefixes across precision, and intervenes directly on every boundary K/V
tensor.  The matched live-cache matrix is closest to the three-pass logic of
\citet{matteson2026refeeding}; the fixed-prefix crossing and bidirectional
transplants ask which state carries the resulting whole-stage trajectory.  We
do not provide their batch-invariant causal closure.

Table~\ref{tab:closestwork} states the overlap and separation directly.  The
novel unit is not re-feeding or precision sensitivity by itself; it is their
combination in a fixed-token, whole-stage task design with live and replay
replica floors, a prospective live/incremental bridge, direct bidirectional KV
transplantation, and an outcome-blind checkpoint replication.

\begin{table}[H]
\centering
\scriptsize
\caption{Design comparison with the two closest studies
\citep{matteson2026refeeding,chodavarapu2026kvcache}.  ``No'' denotes an
absent design element, not a defect relative to a study's stated objective.}
\label{tab:closestwork}
\begin{tabularx}{\linewidth}{@{}L{0.24\linewidth}YYY@{}}
\toprule
Property & Matteson & Chodavarapu--Xu & This work\\
\midrule
Retained decode-time state resumed at a boundary & Yes; block-aligned KV & No; stepwise cache-on/off & Yes\\
Replica floor & Two exact-resume passes & None reported & Both constructions\\
Whole reasoning-stage boundary & No; post-pivot probes & No; ordinary decoding & Yes; merge boundary\\
Fixed identical prefixes across precision & No precision crossing & No matched \(2\times2\) & Yes; 200 prefixes\\
Prospective live/incremental cache bridge & No & No & Yes; 12 rows\\
Direct KV-cache tensor intervention & No & No; residual-stream patching & Yes; bidirectional at boundary, two checkpoints\\
Task endpoint and horizon & Credit estimates; 32-token probes & Trajectory/accuracy; 128-token main runs, 32-token FP32 control & Merge/answer trajectories and paired labels\\
Independent transplant replication & No & No & Later related checkpoint with outcome-blind selection\\
\bottomrule
\end{tabularx}
\end{table}

\subsection{Stage boundaries in structured reasoning}

Self-consistency samples diverse reasoning paths and selects an answer by
marginalizing over them \citep{wang2023selfconsistency}.
Branch--Solve--Merge decomposes a task into parallel subtasks and fuses their
solutions \citep{saha2024bsm}.  Recent systems move beyond answer-only voting
by training an aggregator on concatenated sample sets \citep{qi2025ssa}, recursively
refining populations of reasoning chains \citep{venkatraman2025rsa},
synthesizing solutions from complete reasoning traces
\citep{fadnavis2026beyondconsensus}, or jointly optimizing parallel trace
generation and aggregation \citep{hamid2026spiral}.
\textsc{PaCoRe} coordinates multi-round parallel exploration through message
passing and synthesis \citep{hu2026pacore}, whereas \textsc{ARTS} decouples
generation from verification \citep{hong2026arts}.

For our purposes, these staged decompositions expose intermediate boundaries
at which one can perform ablation, distillation, or counterfactual
measurement.  Counterfactual-likelihood work likewise holds public tokens
fixed while intervening on a private block to measure indirect influence
\citep{lorup2026counterfactual}.  Here the issue is mechanical: a fixed-context
probe can be repeatable while evaluating a fresh-prefill state that the
end-to-end system never occupied.  Replay fidelity is therefore a prerequisite
for interpreting a stage-level rate, not an efficiency detail.

\section{Stage Replay and Experimental Design}

\subsection{Token, role, mask, and replay-boundary contract}
\label{sec:boundarycontract}

For problem \(x\), the model generates three branch traces
\(C=(c_1,c_2,c_3)\), a merge trace \(m\), and an answer \(a\):
\begin{equation}
  C \leftarrow G_\theta(x), \qquad
  m \leftarrow M_\theta(x,C), \qquad
  a \leftarrow A_\theta(x,C,m).
  \label{eq:pipeline}
\end{equation}
The flattened token sequence has exactly six ordered, non-nested blocks:
\begin{align*}
&\texttt{<|prompt|>}\;x\;\texttt{<|/prompt|>}
\;\texttt{<|cot\_1|>}\;c_1\;\texttt{<|/cot\_1|>}\\
&\quad\texttt{<|cot\_2|>}\;c_2\;\texttt{<|/cot\_2|>}
\;\texttt{<|cot\_3|>}\;c_3\;\texttt{<|/cot\_3|>}\\
&\quad\texttt{<|merge|>}\;m\;\texttt{<|/merge|>}
\;\texttt{<|answer|>}\;a\;\texttt{<|/answer|>}.
\end{align*}
Opening and closing delimiters receive the role of their block.  Raw roles
\(0,\ldots,5\) denote prompt, the three branches, merge, and answer; the
adapted checkpoint maps them to appended role-embedding rows \(8,\ldots,13\).
At every layer, the attention mask is the conjunction of the causal mask,
key/query padding validity, and the role-visibility relation in
Table~\ref{tab:rolevisibility}.  In particular, answer tokens cannot attend
directly to branch tokens.

\begin{table}[H]
\centering
\small
\caption{Six-role visibility contract.  Each row lists the key roles visible
to a query role in addition to causal ordering.  Branch \(i\) denotes the
corresponding one of the three branch roles.}
\label{tab:rolevisibility}
\begin{tabularx}{0.88\linewidth}{@{}L{0.22\linewidth}Y@{}}
\toprule
Query role & Visible key roles\\
\midrule
Prompt & Prompt\\
Branch \(i\) & Prompt, branch \(i\)\\
Merge & Prompt, branches 1--3, merge\\
Answer & Prompt, merge, answer\\
\bottomrule
\end{tabularx}
\end{table}

The structural logits mask enforces the block order above.  Inside an open
block, ordinary vocabulary items and only that block's closing delimiter are
eligible structural outputs; between blocks, only the next opening delimiter
is eligible.  A closing delimiter is forced when a branch, merge, or answer
reaches its respective 768, 512, or 64 content-token cap.  Greedy selection
uses temperature 0, top-\(p=1\), and top-\(k=0\).

We study the state immediately after the unique \texttt{<|merge|>} token has
been processed and before the first merge-content token is selected.  Let
\(z\) denote the exact integer-token prefix through that delimiter.  In the
primary batch-of-two duplicate design there is no padding; position IDs and
cache positions enumerate non-padding tokens from zero.  Incremental steps
advance both by one, and the one-shot path receives the same role, position,
visibility, and padding arrays reconstructed from the integer IDs.  The live
path reaches \(z\) through cache updates during branch decoding.  The replay
path discards that cache and reconstructs \(z\) by one-shot prefill before the
common decoder generates merge and answer tokens.

We use three construction names throughout.  A \emph{live} cache is retained
from ordinary autoregressive generation.  An \emph{incremental} cache is
rebuilt from fixed tokens one token at a time under teacher forcing.  A
\emph{prefill} cache is built from those tokens in one forward call.  A
\emph{replica} is a second storage-isolated instance or independent repeat
used only to measure the within-construction floor.  The bridge experiment
tests when live and incremental
caches can be treated as equivalent; the main comparisons never conflate
those labels.

\begin{figure}[H]
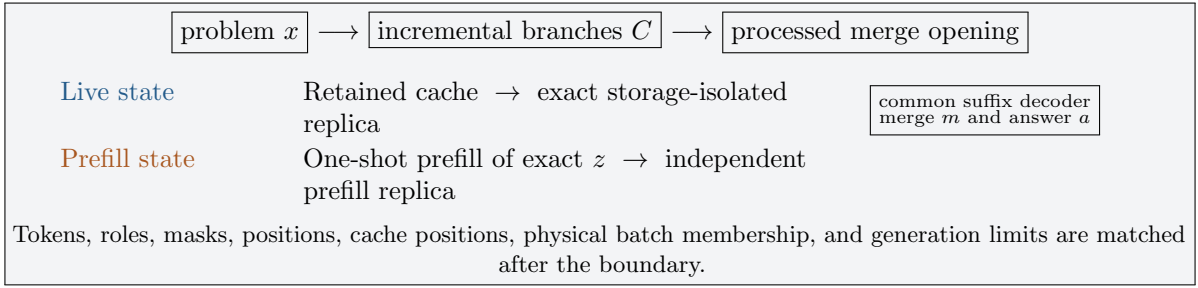

\centering
\fcolorbox{black}{softgray}{%
\begin{minipage}{0.94\linewidth}
\centering
\small
\(\boxed{\text{problem }x}\)
\(\longrightarrow\)
\(\boxed{\text{incremental branches }C}\)
\(\longrightarrow\)
\(\boxed{\text{processed merge opening}}\)

\vspace{0.7em}
\begin{tabularx}{0.92\linewidth}{@{}L{0.18\linewidth}YL{0.23\linewidth}@{}}
\textcolor{routeblue}{Live state} &
Retained cache \(\;\rightarrow\;\) exact storage-isolated replica &
\multirow{2}{*}{\(\boxed{\substack{\text{common suffix decoder}\\
                                   \text{merge }m\text{ and answer }a}}\)}\\
\textcolor{mergeorange}{Prefill state} &
One-shot prefill of exact \(z\) \(\;\rightarrow\;\) independent prefill replica & \\
\end{tabularx}

\vspace{0.5em}
\footnotesize
Tokens, roles, masks, positions, cache positions, physical batch membership,
and generation limits are matched after the boundary.
\end{minipage}}
\caption{Matched live-cache/fresh-prefill design.  The duplicates measure
replica floors.  Conditional on their exactness, the cross-construction
comparison tests whether identical integer tokens reproduce live
continuation.}
\label{fig:replaydesign}
\end{figure}

Live and prefill paths are mathematically intended to represent the same
prefix, but they are not the same computation.  The matched design in
Figure~\ref{fig:replaydesign} holds the suffix surface fixed and changes only
how the boundary state is constructed.

\subsection{Models and evaluation set}

The primary checkpoint is a 14B multi-branch model derived from
Qwen2.5-14B-Instruct \citep{qwen2024qwen25} after a merge-targeted policy
update.  The matched and fixed-prefix experiments use all 200 rows of a frozen
holdout drawn from GPQA Main \citep{rein2024gpqa}, with no exclusions after
decoding.
This domain had been selected after an earlier breadth screen; that screen is
not used as evidence in this paper.  The transplantation replication uses a
distinct later 14B checkpoint from the same training family.  It removes
single-checkpoint dependence but is not an architecture-family replication.

\subsection{Experiment 1: matched live-cache/fresh-prefill matrix}
\label{sec:matchedmatrix}

The experiment generates the boundary once per problem and creates four
storage-isolated continuations:

\begin{itemize}
  \item the retained live \texttt{DynamicCache} and a bit-exact,
        storage-isolated replica;
  \item a one-shot prefill from the same integer-token prefix and a second,
        independently constructed prefill replica.
\end{itemize}

All four arms use one shared suffix decoder and identical token and role
state, visibility masks, positions, cache positions, limits, and physical
batch membership.  Each request contains two duplicate copies of one problem
and fails closed if they desynchronize.  Decoding is greedy through the
Hugging Face Transformers library \citep{wolf2020transformers} and uses
PyTorch SDPA, with 768/512/64 branch, merge, and answer content caps.

The design is run separately in BF16 and FP32.  FP32 disables TF32 matrix
multiplication and serves as a high-precision falsification; neither surface
uses a certified batch-invariant kernel.  An eight-row preflight at each
precision must reproduce both replica pairs exactly.  The full integrity gate
is conjunctive over all 200 problems: both replica pairs must agree exactly in
boundary state, cache tensors, logits, suffixes, and recorded geometry; all
arms must expose the same functional token, role, visibility, position, and
padding state; and every compared cache or logit tensor must have disjoint
storage.

Conditional on those gates, prefill versus live is the primary comparison.
Endpoints are boundary-cache and next-token-logit exactness and magnitude,
complete suffix, merge, and answer disagreement, first divergence position,
paired correctness, and directional correctness flips.  Wilson intervals are
used for binary disagreement rates; accuracy differences use 20,000
problem-paired bootstrap replicates.  The analysis also reports exact
cross-precision boundary-prefix identity.  If BF16 and FP32 reach different
token prefixes, their endpoint difference is interpreted as a
precision-surface comparison rather than a fixed-prefix precision effect.

\subsection{Experiment 2: fixed-prefix precision-by-construction matrix}
\label{sec:fixedprefixdesign}

The decisive precision test freezes one discrete boundary state per problem
before crossing numerical precision.  The bank contains all 200 BF16 token
prefixes from Experiment~1 in source order, without outcome-based filtering;
prefix lengths range from 1,025 to 5,080 tokens.  Every BF16 and FP32 cell
therefore consumes the same integer token, role, visibility, position, and
padding sequence for a given problem.

Within each precision, the incremental cache is constructed by teacher
forcing the fixed prefix one token at a time without allowing the forced
discrete state to change.  It receives a bit-exact, storage-isolated replica.
The prefill cache is built in one call and independently repeated.  The suffix
decoder, physical duplicate batch of two,
generation limits, scoring, and exact integrity gates are otherwise inherited
from Experiment~1.  A two-row preflight at each precision must pass before the
200-row arm runs.

The analysis plan was frozen before either full arm.  The primary estimand is
prefill minus incremental, crossed over precision.  Primary endpoints are
suffix, answer,
and correctness-label disagreement within precision and the problem-paired
FP32-minus-BF16 change in each indicator.  Binary rates use Wilson intervals;
paired differences use 20,000 problem bootstrap replicates, with exact
two-sided McNemar tests as directional side reads.  No rows are excluded.

For incremental and prefill cache tensors \(K_{\mathrm{inc}}\) and
\(K_{\mathrm{pre}}\), relative drift is
\[
  d_{\mathrm{rel}} =
  \frac{\lVert K_{\mathrm{inc}}-K_{\mathrm{pre}}\rVert_2}
       {\max(\lVert K_{\mathrm{inc}}\rVert_2,
             \lVert K_{\mathrm{pre}}\rVert_2)}.
\]
The same quantity is aggregated over all key/value tensors.  Shapes, strides,
dtypes, devices, absolute differences, and raw norms are retained for every
row.

\subsection{Experiment 3: live-to-incremental bridge}
\label{sec:bridgedesign}

Experiment~2 reconstructs its incremental cell by teacher forcing rather
than retaining the original cache from Experiment~1.  We therefore freeze a
prospective bridge before new decode.  Twelve GPQA Main rows are selected by
evenly spaced order statistics after sorting the 200 saved BF16 prefixes by
length; expected answers and prior outcomes are not read.  The selection
spans prefix lengths from 1,025 to 5,080 tokens.

For each row, ordinary greedy BF16 decoding first reaches and retains a new
live merge-boundary cache.  The exact integer-token prefix just reached is
then consumed one token at a time to build an incremental cache in the same
model process.  Each cache receives a bit-exact, storage-isolated replica,
and the live and incremental states are continued with the same physical
duplicate batch and suffix decoder.  Primary endpoints
are cache, boundary-logit, and complete-suffix identity.  Agreement with the
saved prefix bank is descriptive rather than a gate: the experiment
tests construction equivalence conditional on one shared newly reached prefix,
because the earlier cache tensors were not retained.

The saved ledgers also permit a deterministic 200-row behavioral audit without
new decoding.  Joined by problem identifier in frozen order, it compares exact
prefix IDs, complete suffix-token arrays, and every scalar cache/logit summary
shared by the two experiments.  This endpoint can establish trajectory and
numerical-fingerprint reproduction, but not equality of cache tensors whose
values or injective digests were not retained.

\subsection{Experiment 4: bidirectional KV-cache transplantation}
\label{sec:transplantdesign}

We next intervene on the candidate state variable directly.  Before any
transplant outcome is decoded, 32 rows are frozen from the completed BF16
fixed-prefix matrix: 24 of its 166 suffix-divergent rows and eight of its 34
suffix-exact rows.  Within each stratum, rows are evenly spaced
order statistics after sorting by prefix length and problem identifier.  This
outcome-conditioned design tests causal sufficiency on known sensitive states
and includes a bounded procedural control; it does not estimate prevalence.
The selected prefixes span 1,025--5,080 tokens.  Expected answers and
correctness scores are neither read nor recomputed.

Each fixed prefix reconstructs an incremental state and a prefill state in
BF16.  Each construction is instantiated twice with disjoint storage.  The
interventions are bidirectional.  In one direction, all 48 key/value layer
pairs in the prefill recipient are replaced with storage-isolated copies from
the incremental donor; the other direction reverses donor and recipient.  The
recipient's token, role, mask, position, padding, cache position, suffix
decoder, and pre-intervention boundary logits remain unchanged.  The raw
next-token argmax must already agree, so both interventions begin from the
same first suffix token.  For every tensor, the runner verifies provenance,
shape, stride, dtype, sequence length, and absence of storage aliasing.

The registered primary endpoint is exact donor-trajectory recovery in both
directions among rows whose contemporaneous construction baselines diverge.
The negative-control endpoint is whether either full transplant creates a
third trajectory among contemporaneously exact rows.  Binary endpoints use
Wilson 95\% intervals.

\subsection{Experiment 5: outcome-blind checkpoint replication}
\label{sec:transplantrepdesign}

The first transplantation panel is deliberately conditioned on known
divergence.  We therefore freeze a second panel before observing any suffix
or transplant outcome on a later 14B checkpoint from the same training
family.  From the complete 200-prefix BF16 bank, 48 rows are selected as
evenly spaced order statistics after sorting by prefix length and problem
identifier.  Selection reads neither prior suffix identity nor expected
answers or correctness; the selected prefixes span 1,025--5,080 tokens.

Each row reconstructs incremental and prefill states under the later
checkpoint, with the same functional-state, raw-argmax,
storage-isolation, tensor-provenance, and exact-replica gates as
Experiment~4.  Only six arms are permitted: two replicas of each construction
and the two bidirectional whole-cache transplants.  No partial-layer patch or
adaptive refinement is run.  The endpoints are the contemporaneous
incremental/prefill suffix-divergence rate over all 48 rows, donor recovery in each direction and
jointly among divergent rows, and production of a new trajectory among
naturally exact rows.  Wilson 95\% intervals are computed on each eligible
set.

The interpretation matrix is frozen with the design.  ``Strong replication''
requires at least 12 divergent and six exact rows, bidirectional donor
recovery on every divergent row, and no new trajectory among exact controls.
If recovery is complete but fewer than six exact controls occur naturally,
the result is a conditional replication with an underpowered control stratum.
Any recovery failure is checkpoint-sensitive recovery; any novel exact-control
trajectory is a procedural-control failure; and fewer than 12 divergent rows
is divergence-underpowered.

\subsection{Scoring and uncertainty}

Answers are scored with one deterministic matcher that normalizes answer
blocks, multiple-choice forms, and common mathematical formats.  Accuracy
comparisons join arms by problem identifier.  We report paired percentile
bootstrap intervals and exact two-sided McNemar tests on discordant
correctness labels.  McNemar tests describe directional imbalance; a
nonsignificant test does not establish equivalence.

\subsection{Execution contract and reproducibility boundary}
\label{sec:reproducibility}

The primary model is derived from \texttt{Qwen/Qwen2.5-14B-Instruct} and
uses 48 decoder layers with role-conditioned embeddings and the six-role
token/mask contract in Section~\ref{sec:boundarycontract}.  Raw multi-branch
roles 0--5 index appended embedding rows 8--13; visibility is computed from
the raw roles before that offset is applied.  The exact inference identity is:

\begin{center}
\scriptsize
\begin{tabular}{@{}ll@{}}
Weights SHA-256 & \texttt{c63e1ba51169715a558c054478f2bacb991891adb7e75c18512a585af916dfcf}\\
Config SHA-256 & \texttt{dec4e645b4f82cad0a0d7951aa74b4df94333696c19ce5530ec5b8eb7ce1b083}\\
Tokenizer SHA-256 & \texttt{305e7c9ffdfaacde7757cd8b251118a82ffb8809b1be677e5db8289ad3a680bf}\\
Role embedding SHA-256 & \texttt{c8d03cb0a50d7ffaca51d54ef7073e0a63c08911ce0d079ba9e4b05f3dfa24e1}\\
\end{tabular}
\end{center}

These digests distinguish the primary checkpoint from the public backbone.
The outcome-blind transplantation replication uses a distinct later training
checkpoint in the same model family.  Its inference identity is:

\begin{center}
\scriptsize
\begin{tabular}{@{}ll@{}}
Weights SHA-256 & \texttt{78b7eb8eebdc9db7d4adf996cf19febe7cf46ebd8483e5f02ac33eb3272bae6d}\\
Config SHA-256 & \texttt{dec4e645b4f82cad0a0d7951aa74b4df94333696c19ce5530ec5b8eb7ce1b083}\\
Tokenizer SHA-256 & \texttt{366610f4739162cf51637d1291194beb535eab53a861fd3fca0900e254022d29}\\
Role embedding SHA-256 & \texttt{c257d2881a2ec1c0aa9db36c03155ca9be704e5ae1df28c47e4cc6888b6913f4}\\
\end{tabular}
\end{center}

Table~\ref{tab:reprocontract} gives the numerical contract for the matched HF
experiments.  Each \texttt{DynamicCache} contains one key and one value tensor
per layer, each shaped \([2,8,|z|,128]\); the evidence records shape, stride,
dtype, and device for every compared tensor.  The live replica is produced by
a deep copy of the complete cache under inference mode.  The runner rejects a replica if
its tensor inventory, values, metadata, or inference-tensor status changes,
or if any key/value storage aliases the source.  Fresh-prefill replicas are
independent model calls rather than copies.

\begin{table}[H]
\centering
\scriptsize
\renewcommand{\arraystretch}{1.04}
\caption{Reproducibility contract for the primary matched HF matrices.  Values
combine the frozen manifests with a read-only query of the preserved
execution environment.}
\label{tab:reprocontract}
\begin{tabularx}{\linewidth}{@{}L{0.25\linewidth}Y@{}}
\toprule
Field & Frozen value\\
\midrule
Software & Python 3.12.13; PyTorch 2.10.0+cu128; Transformers 5.3.0;
CUDA runtime 12.8; cuDNN 9.10.2; NVIDIA driver 590.48.01\\
Hardware & One NVIDIA B200 (compute capability 10.0), tensor parallelism 1\\
Attention and execution & Hugging Face SDPA; eager inference mode;
\texttt{torch.compile} disabled; deterministic-algorithm mode disabled\\
Numeric flags & BF16: TF32 matrix multiplication disabled and cuDNN TF32
enabled; FP32: both disabled and float32 matmul precision set to
\texttt{highest}\\
Decoding & Greedy (temperature 0, top-(p=1), top-(k=0)); physical batch 2
containing duplicate copies of one problem; branch/merge/answer content caps
768/512/64\\
Boundary state & Exact integer token IDs, role IDs, visibility state,
positions, cache positions, padding state, cache sequence length, and tensor
metadata are checked before continuation\\
Fixed-prefix crossing & The same 200 frozen integer prefixes are used in all
four precision-by-construction cells; incremental caches are teacher-forced
one token at a time; prefix lengths 1,025--5,080\\
Live/incremental bridge & Twelve outcome-blind prefix-length order statistics;
new live and incremental caches constructed in one BF16 process;
saved-prefix agreement retained as an outcome\\
Saved-ledger bridge audit & Exact 200-row problem order and prefix IDs;
complete suffix-token arrays plus all scalar leaves common to both retained
cache/logit comparison schemas; no expected-answer read\\
KV transplantation & Twenty-four prior-divergent and eight prior-exact
length-spread prefixes; bidirectional 48-layer K/V swaps; no correctness
scoring or layer search\\
Outcome-blind transplantation & Forty-eight length-spread prefixes selected
by prefix length and problem identifier only; distinct later checkpoint;
bidirectional 48-layer K/V swaps only; no correctness scoring, partial patch,
or adaptive refinement\\
Cache reporting & Maximum and mean absolute difference, relative \(L_2\)
norm, and tensor shapes, strides, dtypes, and devices retained per row\\
Population and scoring & All 200 rows of the frozen GPQA Main holdout, in
source order; no exclusions; deterministic answer matcher joined by problem
identifier\\
\bottomrule
\end{tabularx}
\end{table}

The reproducibility release is scoped to the claims in this paper.  It
contains a redacted environment/checkpoint manifest, frozen problem-ID order
and dataset digests, exact boundary-prefix and suffix token IDs, per-item
scores, cache-difference summaries, and the
primary generation and analysis harnesses.  It excludes prompt text, private
training corpora, training lineage, unrelated auxiliary experiments, and
internal filesystem names.  Thus the released rows permit independent
recomputation of every reported primary statistic and inspection of tensor
drift; an exact end-to-end rerun still requires access to the
digest-identified adapted checkpoint.

\section{Results}
\label{sec:results}

\subsection{Retained live-cache and fresh-prefill continuations diverge in BF16}

The retained-live four-arm matrix passes every integrity gate at both
precisions.  Both replica pairs are exact on all 200 items, including
boundary caches, logits, decoded suffixes, and correctness labels.  Physical
duplicates remain synchronized, tensor stores are disjoint where required,
and the arms share the same suffix decoder and recorded geometry.  The
live/prefill contrast therefore has no observed replica floor.

In BF16, the two state constructions are non-bit-exact at the boundary on
every item, although their immediate greedy argmax always agrees.  The
continuations subsequently diverge on 166/200 suffixes (83.0\%; Wilson 95\%
CI [77.18,87.57]), 49/200 answers (24.5\%; [19.06,30.90]), and 20/200
correctness labels (10.0\%).  The first suffix mismatch occurs a median of
88 tokens after the boundary (range 4--453), never on the first suffix token.
Live and prefill score 93/200 and 95/200.  The \(+1.0\)-point difference has
a paired 95\% CI of \([-3.5,+5.5]\), with nine live-only and eleven
prefill-only successes (exact McNemar \(p=.824\)).  The result is trajectory
and item churn, not an accuracy gain.

The independently generated FP32 matrix retains small non-bit-exact cache and
logit differences but produces no decoded disagreement.  Because only 1/200
boundary prefixes is token-identical across those original precision runs,
this first matrix establishes live-state failure in BF16 but supplies only a
high-precision surface falsification.

\subsection{A fixed-prefix crossing identifies precision moderation}
\label{sec:fixedprefixresult}

The prospective \(2\times2\) removes that confound.  The integer prefix is
identical across precision on 200/200 problems.  At each precision, both
replica pairs are exact on all 200 boundary caches, logits, functional-state
fields, suffixes, answers, and labels; storage-disjointness and recorded
geometry also pass on every row.

On these fixed states, BF16 again changes 166 suffixes, 49 answers, and 20
correctness labels under one-shot prefill, whereas FP32 changes 0/200 of
each.  Each zero-event rate has Wilson 95\% interval \([0,1.88]\%\).
The paired FP32-minus-BF16 reductions are 83.0 percentage points for suffix
disagreement (95\% bootstrap CI [77.5,88.0]), 24.5 points for answer
disagreement ([18.5,30.51]), and 10.0 points for correctness churn
([6.0,14.5]).  Exact paired tests give \(p=2.14\times10^{-50}\),
\(3.55\times10^{-15}\), and \(1.91\times10^{-6}\), respectively.  BF16
incremental/prefill accuracy is 93/200 versus 95/200; FP32 is 92/200 versus
92/200.  The precision difference in the accuracy effect is \(-1.0\) point
(95\% CI \([-5.5,+3.5]\)), so the result concerns fidelity rather than gain.

\begin{table}[H]
\centering
\scriptsize
\caption{Fixed-prefix precision-by-construction matrix on all 200 items in the
frozen GPQA Main holdout.  Every precision cell uses the same integer prefix.
The incremental and prefill constructions are primary; each has a
storage-isolated replica.  Drift values are medians of per-item summaries.}
\label{tab:matchedcausal}
\begin{tabular}{@{}lrr@{}}
\toprule
Quantity & BF16 & FP32\\
\midrule
Prefix exact across precision & 200/200 & 200/200\\
Both replica pairs exact & 200/200 each & 200/200 each\\
Incremental/prefill cache exact & 0/200 & 0/200\\
Cache maximum absolute difference & 4.409 & 0.001857\\
Cache relative \(L_2\) difference & 0.01224 & \(4.270\times10^{-6}\)\\
Incremental/prefill logits exact & 0/200 & 0/200\\
Logit maximum absolute difference & 0.4375 & 0.0001044\\
Immediate argmax different & 0/200 & 0/200\\
Suffix different & 166/200 & 0/200 (95\% CI [0,1.88]\%)\\
Answer different & 49/200 & 0/200 (95\% CI [0,1.88]\%)\\
Correctness label different & 20/200 & 0/200 (95\% CI [0,1.88]\%)\\
Incremental / prefill correct & 93 / 95 & 92 / 92\\
Prefill-minus-incremental accuracy (95\% CI) & \(+1.0\) pp ([-3.5,+5.5]) & 0.0 pp ([0,0])\\
\bottomrule
\end{tabular}
\end{table}

The cache remains non-bit-exact at both precisions, but its median relative
\(L_2\) difference falls from 0.01224 in BF16 to
\(4.27\times10^{-6}\) in FP32.  Conditional on exact replicas and fixed
integer states, numerical precision therefore moderates whether the
construction difference becomes behaviorally visible.  The experiment does
not isolate a unique low-level cause: reduction order, kernel selection, cache
implementation, and their interactions remain candidates.

\subsection{Incremental construction prospectively reproduces live cache}
\label{sec:bridgeresult}

All 12 bridge rows pass the replica, state, storage, and physical-duplicate
integrity gates.  Conditional on the newly reached live token prefix, the
incremental construction is bit-exact to the retained live
cache on all 48 layers for 12/12 rows.  Boundary logits and complete greedy
suffixes are likewise exact on 12/12.  The newly reached prefixes span
1,463--4,642 tokens; the outcome-blind selection had been spread over
saved-prefix lengths of 1,025--5,080 tokens.

None of the 12 newly decoded prefixes is token-identical to its earlier saved
prefix.  The result therefore does not claim recovery of discarded cache
tensors.  It establishes the relevant prospective bridge: when ordinary
decoding and incremental construction consume the
same newly reached prefix under this HF contract, they produce the same cache,
logits, and continuation.  Experiment~2's incremental cell is consequently
validated as an implementation of live-style cache updates on these tested
states.

The saved-ledger join closes the corresponding behavioral endpoint.  Prefix
IDs and problem order agree on 200/200 rows.  The later incremental arm
reproduces the original live suffix token for token on 200/200 rows; both
replica and prefill arms also reproduce on 200/200.
All three suffix-comparison objects agree on every row.  For each cache
comparison, all 1,158 scalar leaves common to the two schemas agree on 200/200
rows; the analogous 14 common logit leaves agree on 200/200 for each logit
comparison.  These per-tensor drift fingerprints are highly constraining, but
they are not injective hashes of discarded tensors.  The bridge therefore
establishes 12-row prospective tensor equality and 200-row trajectory and
fingerprint equality, not retrospective tensor equality.

\subsection{Divergent continuations follow the transplanted KV cache}
\label{sec:transplantresult}

The transplantation run completes all 32 frozen rows.  Every row passes the
prefix, functional-state, replica, tensor-provenance, storage, and raw-argmax
gates, and all 24 prior-divergent and eight prior-exact strata are
reproduced by the contemporaneous incremental/prefill baselines.

On every divergent row, each complete transplant produces its cache donor's
suffix (Table~\ref{tab:transplant}).  Bidirectional recovery therefore holds on
24/24 selected rows (95\% Wilson CI [86.2,100]\%).  Among the eight
nondivergent controls, neither full-cache direction produces a trajectory
outside the shared baseline (0/8; [0,32.4]\%).  The latter interval reflects
the deliberately small procedural-control stratum and does not establish a
low population-wide artifact rate.

\begin{table}[H]
\centering
\small
\caption{Registered whole-cache transplant endpoints.  Rates are over rows
eligible for the corresponding endpoint.}
\label{tab:transplant}
\begin{tabular}{@{}lrr@{}}
\toprule
Endpoint & Result & Wilson 95\% CI\\
\midrule
Prefill recipient follows incremental-cache donor & 24/24 & [86.2,100]\%\\
Incremental recipient follows prefill-cache donor & 24/24 & [86.2,100]\%\\
Both directions recover their donor & 24/24 & [86.2,100]\%\\
Exact-control row with a third trajectory & 0/8 & [0,32.4]\%\\
\bottomrule
\end{tabular}
\end{table}

The full transplant changes only the cached K/V tensors: recipient boundary
logits and all non-cache functional state are retained, and the immediate
argmax is shared.  The downstream switch must therefore be mediated by the
donated cache once suffix decoding resumes.  On these selected sensitive
states, the complete boundary K/V cache is causally sufficient to select
between the two observed greedy trajectories.  This is stronger than
localization by residual-stream elimination and instantiates the class of
KV-tensor intervention identified as future work by
\citet{chodavarapu2026kvcache}, here at a whole-stage boundary.

\subsection{Outcome-blind transplantation replicates on the later checkpoint}
\label{sec:transplantrepresult}

All 48 frozen rows complete and pass the prefix, functional-state,
raw-argmax, storage-isolation, tensor-provenance, and exact-replica gates.
The later checkpoint's contemporaneous incremental and prefill suffixes
diverge on 43/48 rows (89.6\%; Wilson 95\% CI [77.8,95.5]\%).  This is an
outcome-blind, deterministic length-spread panel rather than a sample chosen
for known sensitivity.

On all 43 divergent rows, each transplant exactly follows its cache donor;
joint bidirectional recovery is therefore 43/43 (95\% CI [91.8,100]\%).  The
other five rows are naturally exact across constructions, and neither
transplant direction creates a trajectory outside
their shared baseline (0/5; [0,43.4]\%).  Table~\ref{tab:transplantrep}
reports the registered endpoints.

\begin{table}[H]
\centering
\small
\caption{Outcome-blind whole-cache replication on the later 14B checkpoint.
Rows were selected by prefix length and problem identifier before any
later-checkpoint suffix or transplant outcome was observed.}
\label{tab:transplantrep}
\begin{tabular}{@{}lrr@{}}
\toprule
Endpoint & Result & Wilson 95\% CI\\
\midrule
Incremental/prefill suffix divergence & 43/48 & [77.8,95.5]\%\\
Prefill recipient follows incremental-cache donor & 43/43 & [91.8,100]\%\\
Incremental recipient follows prefill-cache donor & 43/43 & [91.8,100]\%\\
Both directions recover their donor & 43/43 & [91.8,100]\%\\
Exact-control row with a new trajectory & 0/5 & [0,43.4]\%\\
\bottomrule
\end{tabular}
\end{table}

The frozen interpretation is \emph{conditional replication with an
underpowered control stratum}: donor recovery is complete, but only five
naturally exact controls occur, one fewer than the registered minimum of six
for the strongest label.  The wide 0/5 interval therefore remains explicit.
That limitation concerns the precision of the procedural-control bound, not
the separately conditioned 43/43 recovery endpoint.  Together with the
primary-checkpoint intervention, the result removes known-divergence
selection and single-checkpoint dependence from the donor-recovery finding.
It does not turn the 48 fixed, primary-checkpoint-reachable prefixes into a
model- or task-general prevalence sample.

\section{Implications}
\label{sec:implications}

Replay validation requires four separate statements.  Replicas establish the
floor within each construction.  Exact token and execution contracts establish
that the constructions represent the same discrete prefix.  A retained live
reference tests whether reconstruction preserves the state actually reached.
A direct state transplant tests whether the proposed state variable carries
the divergent trajectory.  The first two statements do not imply the third;
the third does not by itself imply the fourth.

The matched matrix, fixed-prefix crossing, bridge, and transplantation
experiments instantiate those four checks in sequence.  They also show why
aggregate accuracy is an incomplete endpoint: opposing item flips can cancel
while trajectories and answers change.  Fidelity claims should therefore name
the level being compared---cache, suffix, answer, or correctness---and retain
paired item outcomes.

The same requirement applies whenever stored intermediate text substitutes
for a state reached during execution: merge-stage probes, counterfactual token
credit, privileged-context distillation, branch adjudication, and stage-
specific ablations.  Replay remains useful, but until a live-state comparison
passes, it measures behavior from the reconstructed state rather than from the
end-to-end decoder state.

\section{Limitations and Threats to Validity}

\paragraph{Mechanistic resolution.}
The matched experiment isolates the implemented live-cache versus
fresh-prefill construction under one shared suffix surface.  The fixed-prefix
crossing establishes precision moderation on the frozen states, but changing
precision changes all arithmetic in both cache construction and suffix
decoding.  It therefore supports precision moderation but does not identify a
unique kernel operation or distinguish reduction order from other numerical
path differences.  Neither precision arm uses a certified batch-invariant
kernel.

The prospective bridge is tensor-exact on 12 newly reached prefixes, but none
matches an earlier saved prefix and the earlier cache tensors were not
retained.  The 200-row audit establishes trajectory and comparison-fingerprint
reproduction, not retrospective tensor equality.

The transplantation experiments identify the complete K/V state as a
causally sufficient carrier on their divergent boundaries, not as the unique
numerical origin of the drift.  They swap keys and values jointly and
therefore do not separate their roles or identify one layer or operation.

\paragraph{External validity.}
The two checkpoints belong to one Qwen2.5-derived multi-branch family.  The
matched matrix uses one frozen 200-item holdout drawn from GPQA Main, one
B200, Hugging Face SDPA, greedy decoding, and physical duplicate batches of
two.  The later-
checkpoint transplant removes single-checkpoint dependence but not
architecture-, runtime-, hardware-, task-, or sampling-temperature
dependence.

An adjacent stock-model diagnostic offers one bounded cross-check.  vLLM
documents its batch-invariance mode as a beta feature intended to make output
independent of batch size and request order \citep{vllm2026batchinvariance}.
With the public Qwen2.5-14B-Instruct checkpoint under vLLM 0.24.0 on a B200,
and with that mode attested as enabled, physical-pair replicas are exact in
tokens and selected-token log probabilities on 10/10 prompts; singleton
versus paired execution preserves the token trajectory on 7/10 and the
selected-log-probability trace on 6/10.  The beta documentation lists
Qwen2.5-14B-Instruct among its tested and verified models
\citep{vllm2026batchinvariance}.
Thus batch-composition sensitivity occurs without the custom role embeddings,
visibility mask, structural processor, or stage scheduler on this runtime.
Because the diagnostic changes batch composition rather than live/prefill
cache construction, it is not a stock-model stage-replay replication and does
not broaden the main claim.  Frozen artifacts retain checkpoint and runtime
attestations plus per-prompt traces.

\paragraph{Population and selection.}
GPQA Main was selected after an earlier breadth screen, although every row in
the frozen 200-item stratum enters the matched and fixed-prefix analyses.  The
prefix bank is a BF16-reachable state population rather than an architecture-
neutral sample.
The transplant subset is explicitly conditioned on prior suffix divergence:
24 length-spread sensitive rows plus eight length-spread exact controls.  Its
24/24 recovery rate estimates sufficiency within that selected stratum, not
the frequency of cache-mediated divergence in the full population; the 0/8
control result has a 32.4\% Wilson upper bound.  The later 48-row replication
does not use prior suffix outcomes, but its deterministic length-spread panel
is drawn from the same primary-checkpoint-reachable prefix bank.  Its 43/48
divergence rate is therefore a panel-specific estimate, and only five rows are
naturally exact; the 0/5 control result has a 43.4\% upper bound and misses the
registered six-control threshold by one.

\paragraph{Semantic and scoring validity.}
Automated answer matching is imperfect, although suffix-token and cache
endpoints do not depend on that matcher.  Correctness churn and accuracy are
therefore secondary to the exact trajectory and cache endpoints.

\section{Conclusion}

Stage replay can be exactly repeatable and still fail to reproduce live
execution.  The matched and fixed-prefix matrices separate replica stability
from state fidelity: one-shot prefill of identical tokens changes BF16
trajectories despite exact within-construction replicas, while the tested FP32
states show no decoded disagreement.  Prospective live/incremental equality
and the saved-ledger audit connect incremental construction to live
decoding without claiming equality for discarded tensors.

Direct intervention identifies the complete boundary K/V cache as a causally
sufficient carrier on the tested divergent states.  Bidirectional donor
following replicates at a later checkpoint under outcome-blind selection.
The practical conclusion is a measurement rule: a reconstructed context
should not be interpreted as the state an end-to-end decoder occupied until
the comparison includes a retained live reference, replica floors, an exact
token/role/mask/position contract, and the endpoint of interest.

\section*{Acknowledgements}

Large language model tools were used for language editing and drafting
assistance, as the author is not a native English speaker.  All experimental
design, implementation, analysis, and scientific claims are the author's own,
and all reported results are independently recomputable from the released
artifacts.

\begingroup
\sloppy
\hbadness=2000
\bibliographystyle{plainnat}
\bibliography{stage_replay_repeatable_precision_sensitive}
\endgroup

\end{document}